\documentclass[10pt]{article} % For LaTeX2e

\usepackage[preprint]{rlj} % Should be uncommented for preprint versions

\usepackage{amssymb}            % Defines common symbols like \mathbb R
\usepackage{mathtools}          % Extends amsmath, providing common math tools
\usepackage{mathrsfs}           % Enables \mathscr, which can work in cases that \mathcal does not
\usepackage{graphicx}           % For including images
\usepackage{subcaption}         % Allows for the use of subfigures and subcaptions
\usepackage[space]{grffile}     % For spaces in image names
\usepackage{url}                % For displaying URLs
\usepackage{lipsum}             % For placeholder text

\usepackage{adjustbox}
\usepackage{tikz}
\usetikzlibrary{arrows.meta, positioning, shapes.geometric}

\usepackage{booktabs}
\usepackage{multirow}
\usepackage{tabularx}
\usepackage[percent]{overpic}

\title{Towards Scalable Multi-Task Reinforcement Learning with Large Decision Models}

\setrunningtitle{Large Decision Model}

\author{Thibaut Kulak\textsuperscript{1}}

\emails{thibaut.kulak@neoinstinct.com}

\affiliations{
\textsuperscript{1}\textbf{NeoInstinct SA}\\
}

\begin{document}

% \makeCover  % Create the cover page
\maketitle  % Make the title section

% \begin{abstract}
% Inspired by last years' progress on large-scale language processing, we build a generalist agent for quantitative decision-making. The agent, which we call LDM-v0, is a generalist multi-modal and multi-task Reinforcement Learning policy conditioned on past observations/actions/rewards and a given observation. In this paper we describe the data and method for training LDM-v0, and its current capabilities. 
% \end{abstract}

\begin{abstract}
Recent progress in large-scale sequence modeling has shown that a single model can learn useful representations across highly diverse data distributions. Inspired by these advances, we investigate whether a unified transformer policy can be trained across large collections of heterogeneous reinforcement learning environments.

We introduce LDM-v0, a Large Decision Model trained offline on trajectories collected from thousands of environments spanning multiple domains and modalities. LDM-v0 is a multi-task, multi-modal transformer policy conditioned on histories of observations, actions, rewards, and termination signals, and trained through supervised next-action prediction over offline trajectories. We describe the environment infrastructure, automated data generation pipeline, model architecture, and training methodology used to build LDM-v0, and evaluate its performance across diverse environments. We show that a single pretrained model matches the performance of independently trained task-specific reference policies on approximately 1,000 environments including robotics, autonomous driving, inventory management, cybersecurity, trading, and video games. These results demonstrate the feasibility of large-scale offline pretraining across heterogeneous reinforcement learning environments using a single transformer policy.

\end{abstract}

%%%%%%%%%%%%%%%%%%%%%%%%%%%%%%%%%%%%%%%%%%%%%%%%%%%%%%%%%%%%%%%%
%% Section: Submission of papers to RLJ/RLC
%%%%%%%%%%%%%%%%%%%%%%%%%%%%%%%%%%%%%%%%%%%%%%%%%%%%%%%%%%%%%%%%

\section{Introduction}
\label{sec:intro}

Reinforcement Learning (RL) provides a general framework for sequential decision making and has achieved impressive results in domains such as games, robotics, resource optimization, and control. Despite this progress, applying RL in real-world settings remains difficult. Modern RL systems often require extensive environment interaction, careful reward engineering, domain-specific architectures, and substantial hyperparameter tuning. As a consequence, many successful applications rely heavily on expert knowledge and task-specific design choices.

Offline RL and offline-to-online RL partially address these limitations by leveraging previously collected trajectories to reduce costly online interaction. However, selecting and adapting a suitable RL algorithm for a new environment remains challenging \citep{nie2022data}. These difficulties motivate the development of more general and automated approaches to reinforcement learning.

In parallel, large-scale sequence models trained on diverse datasets have transformed natural language processing and, more recently, computer vision and multimodal learning. In RL, sequence-modeling approaches such as Decision Transformers \citep{chen2021decision} have shown that policies can be represented as autoregressive models over trajectories. These results raise an important question: can multi-domain offline RL trajectories be consolidated into a single scalable transformer policy while maintaining strong task performance across many domains? One key challenge is that multi-domain RL ecosystems are fragmented and hard to unify.

In this work, we explore this direction by building a unified multi-domain RL infrastructure, using it to generate large-scale trajectories and training LDM-v0, a Large Decision Model instantiated as a single transformer policy. LDM-v0 is a multi-task and multi-modal model conditioned on past observations, actions, rewards, and current observations in order to predict future actions. Our primary objective is not to study out-of-distribution generalization, but rather to investigate whether a single transformer policy can jointly model diverse RL behaviors at scale.

We present the environment infrastructure, large-scale RL dataset generation pipeline, model architecture and training methodology used to train LDM-v0, and evaluate its performance across a diverse collection of environments. More broadly, we view LDM-v0 as a step toward scalable pretrained reinforcement learning systems.% trained through unified offline pipelines.

%%%%%%%%%%%%%%%%%%%%%%%%%%%%%%%%%%%%%%%%%%%%%%%%%%%%%%%%%%%%%%%%
%% Section: Citations, figures, tables, references, equations
%%%%%%%%%%%%%%%%%%%%%%%%%%%%%%%%%%%%%%%%%%%%%%%%%%%%%%%%%%%%%%%%
\section{Related Work}
\label{sec:rw}

Transformer architectures \citep{vaswani2017attention} have recently become an important framework for reinforcement learning, motivated by their success in large-scale sequence modeling \citep{brown2020language}. Sequence-modeling approaches such as Decision Transformer \citep{chen2021decision} and Trajectory Transformer \citep{janner2021offline} demonstrated that RL policies can be formulated as autoregressive models over trajectories, while subsequent work has further explored the role of transformers in RL \citep{agarwal2023transformers}.

Our work is also related to meta-reinforcement learning. Traditional meta-RL methods \citep{finn2017model, duan2016rl, beck2023survey} aim to learn agents that can adapt rapidly across tasks, often through explicit task distributions, recurrent policies, or gradient-based adaptation. More recently, in-context RL approaches have investigated whether transformers can learn adaptation strategies directly from trajectory context \citep{laskin2022context, lee2023supervised, team2023human, grigsby2023amago, grigsby2024amago, kumarlearning, sridhar2024regent, petrov2024transformers, raparthy2023generalization}. In particular, \citet{lee2023supervised} cast meta-RL as supervised pretraining, where a transformer predicts actions conditioned on a query state and an in-context dataset of prior interactions. This perspective has also been connected theoretically to Bayesian posterior sampling \citep{wang2024understanding}.
% More recently, several works explored the connection between transformers, meta-reinforcement learning, and in-context adaptation \citep{laskin2022context, lee2023supervised, team2023human, grigsby2023amago, grigsby2024amago, kumarlearning, sridhar2024regent, petrov2024transformers, raparthy2023generalization}. These approaches suggest that sufficiently large sequence models can leverage trajectory context to improve decision making.

LDM-v0 is most closely inspired by Gato \citep{reed2022generalist} and supervised in-context RL pretraining \citep{lee2023supervised}. Unlike prior generalist transformer agents \citep{reed2022generalist, gallouedec2024jack} that combine RL with language, robotics, or internet-scale supervised data, LDM-v0 focuses specifically on scalable reinforcement learning pretraining across highly heterogeneous RL ecosystems. Our work emphasizes multi-domain environment integration (see Table \ref{tab:related_work_scale}), large-scale trajectory generation across thousands of Gym/Gymnasium-compatible environments, and compact transition-level sequence modeling.

\begin{table}[t]
\centering
\small
\setlength{\tabcolsep}{6pt}
\renewcommand{\arraystretch}{1.12}
\begin{tabular}{lcccc}
\toprule
\multirow{2}{*}{Model} 
    & \multicolumn{2}{c}{Training scale} 
    & \multirow{2}{*}{Modalities} 
    & \multirow{2}{*}{Tokens / timestep} \\
\cmidrule(lr){2-3}
    & RL libraries 
    & Tasks / envs 
    &  
    &  \\
\midrule
Gato \citep{reed2022generalist} 
    & 16 
    & 596 
    & Multi-modal 
    & Variable, per-dimension \\

JAT \citep{gallouedec2024jack} 
    & 4 
    & $\sim$150 
    & Multi-modal 
    & 2 \\

LDM-v0 (ours) 
    & 146 
    & $\sim$3,000 
    & Multi-modal 
    & 1 \\
\bottomrule
\end{tabular}
\caption{
Comparison of large-scale transformer policies trained across multi-domain reinforcement learning environments. 
The table reports the number of reinforcement learning libraries used, the approximate number of tasks or environments covered, the input/output modalities considered, and the number of tokens used to encode each trajectory timestep. For Gato, this number varies with the dimensionality of the observation and action spaces because different dimensions are tokenized separately; the paper reports more than 100 tokens for a single image observation.
}
\label{tab:related_work_scale}
\end{table}

%%%%%%%%%%%%%%%%%%%%%%%%%%%%%%%%%%%%%%%%%%%%%%%%%%%%%%%%%%%%%%%%
%% Section: Final instructions
%%%%%%%%%%%%%%%%%%%%%%%%%%%%%%%%%%%%%%%%%%%%%%%%%%%%%%%%%%%%%%%%
\section{Method}
\label{sec:method}
% The guiding design principle of LDM-v0 is to be able to train on the widest variety of RL environments. An overview of the model architecture is given in Fig.\ref{fig:architecture}. 

The goal of LDM-v0 is to train a single policy model across a large and heterogeneous collection of reinforcement learning environments. Our approach combines automated reference-policy supervision with a unified sequence-modeling architecture: we first generate supervised policy data by training task-specific RL agents and retaining high-performing policies as references, then train a transformer policy to predict reference actions from interaction histories and current observations.

\subsection{Automated Reference-Policy Supervision}
\label{subsec:automated_ref_policy}

% Generating high-quality trajectories across diverse environments is challenging because no single RL algorithm or hyperparameter configuration performs well across all domains. This creates an AutoRL problem at the scale of dataset construction: for each new environment, one must identify which algorithmic configuration can produce a strong reference policy. Rather than relying on environment-specific expert knowledge, we address this problem with an automated reference-policy pipeline that performs empirical algorithm selection and configuration ranking at scale.

Generating high-quality trajectories across diverse environments is challenging because no single RL algorithm or hyperparameter configuration performs well across all domains. We therefore treat dataset construction as an AutoRL problem: for each environment family, the pipeline empirically ranks candidate algorithms and configurations, trains strong task-specific reference policies, records their trajectories, and annotates collected observations with final-policy actions. LDM-v0 is thus trained to imitate strong task-specific policies rather than exploratory actions. Low-quality runs are removed using performance-based filtering; implementation details are given in Section~\ref{sec:experimental_setup}.

% For each environment family, multiple reinforcement learning algorithms and hyperparameter configurations are evaluated, and the strongest configurations are used to train task-specific reference policies. The resulting interaction trajectories are recorded during training. After training, the final policy for each task is replayed over the collected observations to annotate them with reference actions. Consequently, LDM-v0 is trained to imitate strong task-specific policies rather than the exploratory actions produced during reinforcement learning.

% Finally, low-quality runs are removed using performance-based filtering before constructing the final training dataset. Implementation details of this pipeline, including algorithm budgets, filtering thresholds, and dataset statistics, are provided in Section~\ref{sec:experimental_setup}.

\subsection{LDM-v0 Architecture}

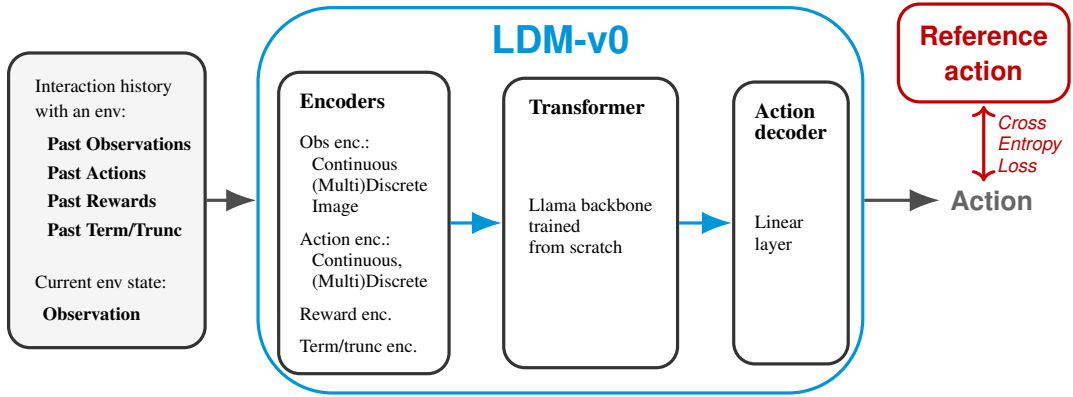
\begin{figure}[t]
% \centering
% \resizebox{\textwidth}{!}{
\begin{tikzpicture}[
    font=\scriptsize,
    box/.style={draw=black!80, very thick, rounded corners=8pt, align=left, inner sep=8pt},
    bigbox/.style={draw=cyan!80!blue, very thick, rounded corners=28pt},
    arrow/.style={-{Latex[length=4mm]}, very thick, draw=black!70},
    bluearrow/.style={-{Latex[length=4mm]}, very thick, draw=cyan!80!blue},
    lossarrow/.style={<->, very thick, draw=red!80!black},
]

% Left context box
\node[box, minimum width=2.6cm, minimum height=3.9cm, fill=gray!8] (context) at (0,0.15) {};
\node[align=left, anchor=north west] at ([xshift=0.25cm,yshift=-0.25cm]context.north west) {
    % \\[0.25cm]
    Interaction history\\[0.05cm]
    with an env:\\[0.15cm]
    \hspace{0.1cm} \textbf{Past Observations}\\[0.1cm]
    \hspace{0.1cm} \textbf{Past Actions}\\[0.1cm]
    \hspace{0.1cm} \textbf{Past Rewards}\\[0.1cm]
    \hspace{0.1cm} \textbf{Past Term/Trunc}\\[0.4cm]
    % \centering $+$\\[0.6cm]
    Current env state:\\[0.15cm]
    \hspace{0.1cm}\textbf{Observation}
};

% LDM container
\node[bigbox, minimum width=8.0cm, minimum height=5.1cm] (ldm) at (6.0,0.15) {};
\node[cyan!80!blue, font=\sffamily\bfseries\Large] at ([yshift=-0.45cm]ldm.north) {LDM-v0};

% Internal boxes
\node[box, minimum width=0.5cm, minimum height=3.5cm] (enc) at (3.4,-0.15) {
    \textbf{\footnotesize Encoders}\\[0.25cm]
    Obs enc.:\\
    \hspace{0.1cm} Continuous\\
    \hspace{0.1cm} (Multi)Discrete\\
    \hspace{0.1cm} Image\\[0.15cm]
    Action enc.:\\
    \hspace{0.1cm} Continuous,\\
    \hspace{0.1cm} (Multi)Discrete\\[0.15cm]
    Reward enc.\\[0.15cm]
    Term/trunc enc.
};

\node[box, minimum width=2.3cm, minimum height=3.5cm] (tr) at (6.4,-0.15) {
    \textbf{\footnotesize Transformer}\\[1.0cm]
    Llama backbone\\
    trained\\
    from scratch \\[0.95cm]
};

\node[box, minimum width=1.2cm, minimum height=3.5cm] (dec) at (9.05,-0.15) {
    \textbf{\footnotesize Action}\\
    \textbf{\footnotesize decoder}\\[0.9cm]
    Linear\\
    layer\\[0.95cm]
};

% Arrows
\draw[arrow] (context.east) -- (ldm.west);
\draw[bluearrow] (enc.east) -- (tr.west);
\draw[bluearrow] (tr.east) -- (dec.west);
\draw[arrow] (ldm.east) -- ++(1.0,0) node[right, gray!80!black, font=\sffamily\bfseries\normalsize] {Action};

% Expert action and loss
\node[box, align=center, red!75!black, font=\sffamily\bfseries\normalsize] (expert) at (11.6,2.1) {
    Reference\\[0.05cm]
    action
};

\node[align=left, red!75!black, font=\sffamily\itshape\scriptsize] at (12.2,0.9) {Cross\\Entropy\\Loss};
\draw[lossarrow] (11.6,0.4) -- (11.6,1.4);

\end{tikzpicture}

\caption{\textbf{Architecture of LDM-v0.} LDM-v0 receives an interaction history and the current observation, encodes each modality, merges them into transition-level embeddings (containing an observation at timestep t and action/reward/done at timestep t-1), processes them with a Llama backbone, and decodes the predicted action. During training, the prediction is supervised using strong task-specific reference policies.}
\label{fig:architecture}
\end{figure}

The primary design objective of LDM-v0 is to support unified training across varied reinforcement learning environments, including domains with different observation modalities, action spaces, and temporal dynamics. An overview of the architecture is shown in Fig.~\ref{fig:architecture}.

LDM-v0 receives an interaction history and the current observation as input. Observations, previous actions, rewards, and termination signals are encoded using modality-specific encoders, merged into transition-level embeddings, processed by a decoder-only transformer backbone, and decoded into an action prediction.

% \begin{center}
% \begin{figure}[h]
% \centering
% \includegraphics[width=0.9\textwidth]{images/architecture.png}
% \caption{\textbf{Architecture of LDM-v0} The inputs to LDM-v0 are a sequence of observations/actions/rewards/dones transitions and a given observation. Those inputs are first encoded using modality-specific encoders, then merged into transition-level embeddings (containing an observation at timestep t and action/reward/done at timestep t-1). Those embeddings are then fed to a transformer backbone that outputs values for each observation in the input sequence. Eventually, the outputs of the transformer are decoded into actions.}
% \label{fig:architecture}
% \end{figure}
% \end{center}

\subsection{Tokenization and Embedding}

LDM-v0 converts multi-modal environment interactions into a unified tokenized representation compatible with transformer sequence modeling.

\begin{itemize}
    \item Continuous inputs are $\mu$-law encoded, discretized into 1024 bins, and mapped through learned embedding tables.
    
    \item Discrete inputs are embedded through lookup tables.
    
    \item Image observations are resized to a common resolution $(64,64,3)$ and encoded using a convolutional encoder.
\end{itemize}

The different dimensions of each observation are stacked into an observation embedding, using a maximum observation dimension of 128 (padding is applied for smaller observations). The same is done for multi-dimensional actions, with a maximum action dimension of 28.
The observation/action/reward/termination embeddings are then aligned (observation aligns with previous action/reward/termination signal; padded in case of first observation), stacked and processed to a transition-level meta-token using a linear layer.

This representation maps multi-modal environment interactions into a shared latent sequence representation compatible with a single transformer backbone. This compact transition-level packing also reduces sequence length compared to per-dimension tokenization approaches \citep{reed2022generalist} and enables longer interaction histories within a fixed transformer context budget.

\subsection{Backbone}
LDM-v0 uses a decoder-only transformer backbone based on the Llama architecture. The backbone processes transition-level embeddings autoregressively and produces contextualized representations used for action prediction.

The model is trained entirely from scratch on the dataset described in Section~\ref{sec:experimental_setup}, as we did not observe measurable improvements from initializing with language-model checkpoints in preliminary experiments.

\subsection{Action Decoder and Training Objective}
The transformer output is passed to a linear action decoder that predicts logits over discretized action bins for each action dimension. For continuous actions, the decoder predicts the corresponding discretized bin; for discrete actions, it predicts the corresponding action category.
% (of shape ($\text{number of bins}) \times (\text{maximal number of actions})$

LDM-v0 is trained to predict reference actions from trajectories generated by RL agents trained independently on each environment (described in Section~\ref{section:datagen}). Training is performed using a standard cross-entropy loss over discretized actions.

Formally, the model learns an autoregressive policy of the form:

$$
a_t = \mathrm{LDM} \Big( (o_i)_{i=1}^{T}, (a_i, r_i, d_i)_{i=1}^{T-1} \Big),
$$

where $o_i$, $a_i$, $r_i$, and $d_i$ respectively denote observations, actions, rewards, and termination indicators. The model autoregressively predicts actions conditioned on the trajectory history retained within the context window.

\section{Experimental Setup}
\label{sec:experimental_setup}

\subsection{Environments}

% Many publicly available RL environments were developed to study practical control, optimization, or sequential decision-making problems and therefore naturally span diverse observation modalities, action spaces, temporal horizons, and reward structures. Since we seek to investigate whether large-scale training across a broad range of existing RL environments is feasible, they represent an interesting source of diverse environments. 

% Our training environments are therefore collected from publicly available GitHub repositories implementing RL environments compatible with the OpenAI Gym or Gymnasium interfaces. These repositories typically depend on different Python versions, package ecosystems, and simulator dependencies, making unified large-scale training difficult in practice.

Public RL environments provide a natural testbed for large-scale heterogeneous policy training: they cover a wide range of control, optimization, and sequential decision-making problems, and vary substantially in observation modalities, action spaces, temporal horizons, and reward structures. We collect training environments from publicly available GitHub repositories implementing OpenAI Gym- or Gymnasium-compatible interfaces. Although these environments share a common high-level API, they often rely on different Python versions, package ecosystems, and simulator dependencies, which makes unified large-scale training difficult in practice.

To address this challenge, we developed an internal environment orchestration framework that encapsulates each environment library within isolated Docker containers and exposes a unified interaction API compatible with modern Gymnasium interfaces. This infrastructure enables scalable and reproducible interaction with multi-domain RL environments while preserving compatibility with legacy dependencies.

The process of fetching, validating, containerizing, and integrating environment repositories into the framework was partly automated. Using this pipeline, we collected 146 environment libraries corresponding to approximately 15,000 individual environments.

The list of integrated libraries and their corresponding number of environments is summarized in Appendix~\ref{appendix:libs}. We note that the number of environments alone is not necessarily indicative of behavioral diversity, as some libraries contain many closely related tasks while others expose fewer but highly configurable environments.

\subsection{Reference-Policy Data Generation}
\label{section:datagen}

We instantiate the automated reference-policy supervision pipeline described in Section~\ref{subsec:automated_ref_policy} as follows. 

We define a fixed pool of candidate algorithm/configuration pairs drawn from Stable-Baselines3 and SB3-Contrib, including A2C, ARS, DDPG, DQN, PPO, QR-DQN, SAC, TD3, TQC, and TRPO. Candidates use either default library hyperparameters or predefined alternatives; the complete set is reported in Appendix~\ref{appendix:rl_candidate_configs}. A candidate is evaluated only when compatible with the environment observation and action spaces, and no hyperparameters are tuned manually for individual environments.

For each environment library, compatible candidates are first trained on a subset of five environments for 3 million transitions. Each run is evaluated by the mean episode return over the last 10\% of training episodes. Candidate configurations are then ranked using pairwise comparisons: for each environment, a configuration receives one point over another only when its final performance is significantly higher according to a one-tailed Welch's t-test. Scores are summed across the five-environment subset to obtain a library-level ranking.

We then select the top-$N$ algorithm/configuration pairs for each environment ($N=3$ in our experiments), and train each for 3 million transitions. Due to computational constraints, we cap the number of environments to 250 per library. During training, all interaction trajectories are recorded, including observations, actions, rewards, and termination signals.

After training, the final trained policy is replayed over all collected observations to generate reference action annotations. Consequently, the supervision targets used to train LDM-v0 correspond to actions produced by the final policy rather than the potentially exploratory actions originally taken during data collection.

We apply several data curation steps to improve dataset quality:

\begin{itemize}
    \item We remove trajectories from training runs that do not exhibit statistically significant performance improvement during learning\footnote{We use a one-tailed Welch's t-test on episode returns from the first and last 10\% of training episodes}.
    
    \item We remove trajectories whose final policy performance falls below 95\% of the best-performing policy trained on the same environment.
\end{itemize}

The resulting dataset contains approximately 4,000 high-performing agents spanning 3,000 distinct environments and a total of 9.3 billion transitions annotated with reference policy actions. A summary of the dataset is provided in Appendix~\ref{appendix:datasets}.

\subsection{LDM-v0 Training Details}

Unless otherwise specified, the main LDM-v0 model has 12 hidden layers, 12 attention heads, a hidden size of 768 and a context length of 2048 transitions. The model has 308M parameters and is trained from scratch for six days using two nodes equipped with eight NVIDIA H200 GPUs each.

We use the AdamW optimizer with $\beta_1=0.9$, $\beta_2=0.999$, $\epsilon=10^{-8}$ and a constant learning rate of $2.5 \times 10^{-4}$. Training uses Distributed Data Parallel with a total batch size of 1024.

The wall-clock time required to train one reference policy for 3 million transitions varies substantially across environments, since the dominant cost is often environment simulation rather than policy optimization. Reference-policy data generation took 12 weeks in total on four servers, consisting of two nodes with 8 NVIDIA H200 GPUs each and two nodes with 4 NVIDIA RTX 4090 GPUs each, for a total of 608 CPU cores. The generated dataset occupies 29 terabytes of storage.

\subsection{Evaluation Protocol} 
For each environment, LDM-v0 performance is averaged over two fresh rollouts of 10 episodes each. The context is initialized empty before each rollout, and the interaction history is kept across episodes during deployment. We use deterministic decoding at evaluation time. The action-space mask is applied to ensure that predicted actions are feasible, for example within the valid bounds for continuous actions or within the allowed range for discrete and multi-discrete actions. We then select the action corresponding to the maximum logit without sampling. Each environment score is normalized relative to the corresponding task-specific reference-policy training trajectory. The average episode return over the first 10\% of the trajectory is treated as poor performance, corresponding to 0\%, while the average over the last 10\% is treated as reference performance, corresponding to 100\%.

\section{Results}
\label{sec:results}
We evaluate whether a single pretrained LDM-v0 model can achieve strong performance across heterogeneous training environments, and study how performance changes with model size.

\subsection{Training Environments}

\begin{figure}[h]
\centering
\begin{overpic}[width=0.95\textwidth]{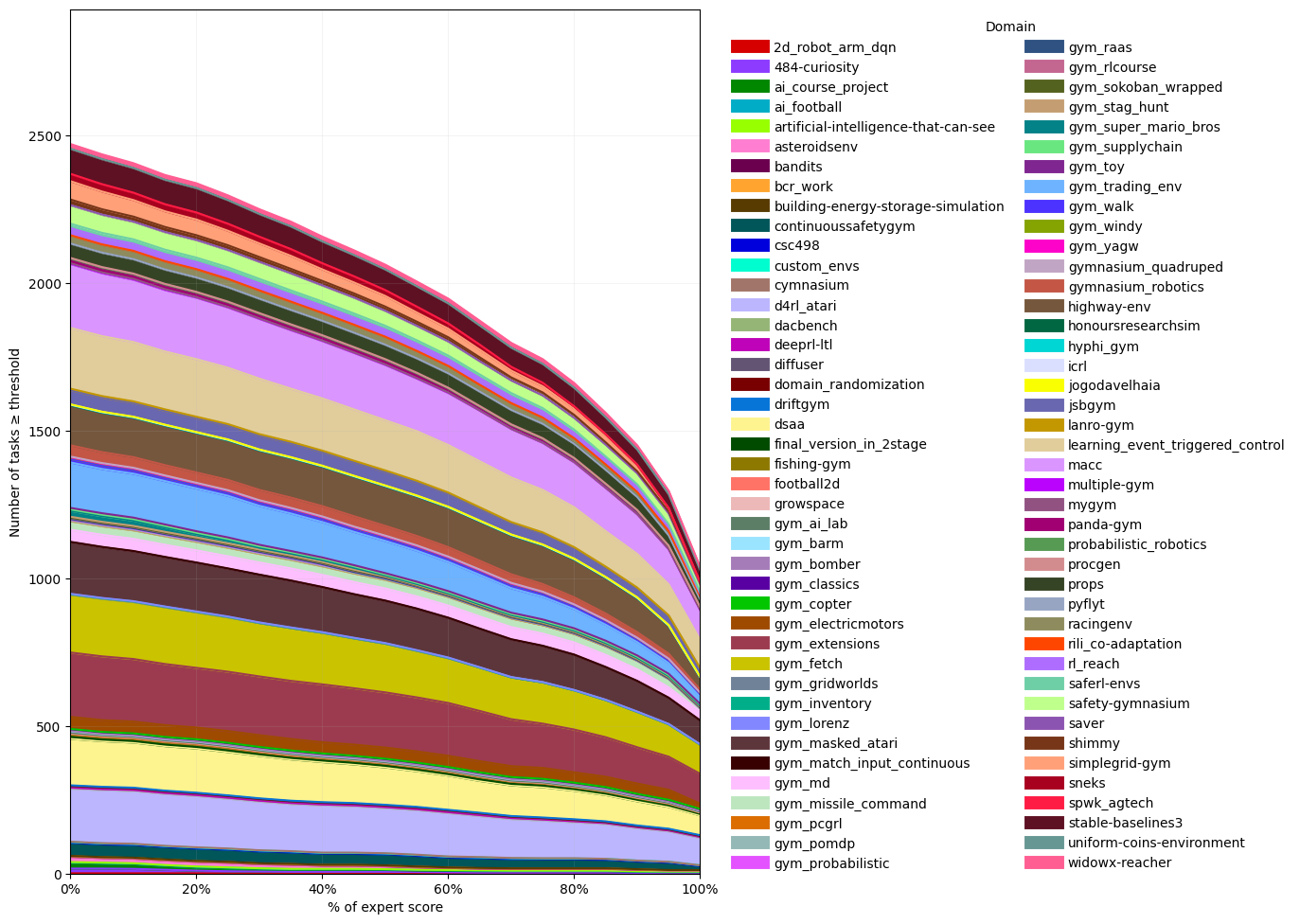}
    \put(32,52){\small \textbf{$>1{,}600$ envs at 80\%}}
    \put(44.2,50){\vector(0,-1){8}}
\end{overpic}
\caption{
\textbf{Performance-threshold curve of LDM-v0 across training environments.}
For each threshold on the x-axis, the y-axis reports the number of environments where LDM-v0 achieves at least that percentage of the corresponding task-specific reference-policy return. 
%The value of the curve at 80\% therefore directly supports the claim that LDM-v0 reaches at least 80\% of reference-policy performance on more than 1,600 environments.
}
\label{fig:trainingresults}
\end{figure}

Figure~\ref{fig:trainingresults} shows a performance-threshold curve for LDM-v0 against task-specific reference agents on training environments. All reported results are obtained using a single pretrained model with a single shared set of parameters across all environments. 

Performance varies substantially across libraries, but strong results are observed across many unrelated domains despite the use of a single shared parameterization. LDM-v0 achieves more than 80\% of the reference policy performance on over 1,600 environments spanning a broad range of domains and matches reference-policy performance on approximately 1,000 environments.

In particular, LDM-v0 demonstrates strong performance across a wide variety of practical sequential decision-making problems. Performance comparable to task-specific high-performing agents is observed in robotic manipulation and control (\texttt{gymnasium\_robotics}, \texttt{panda-gym}), drone and UAV control (\texttt{gym-copter}, \texttt{jsbgym}), autonomous driving simulation (\texttt{highway-env}), electric motor control (\texttt{gym\_electricmotors}), smart-grid and energy-management tasks (\texttt{building-energy-storage-simulation}), financial trading (\texttt{gym\_trading\_env}), inventory optimization (\texttt{gym\_inventory}), cybersecurity-oriented environments (\texttt{cymnasium}), and plant or crop optimization tasks involving branching and growth control (\texttt{growspace}).

Beyond real-world-inspired domains, LDM-v0 also achieves strong results on complex game-like environments including Atari-style tasks (\texttt{gym\_masked\_atari}, \texttt{gym\_super\_mario\_bros}), and procedurally generated environments such as \texttt{procgen}.

% The ability of a single set of parameters to achieve competitive performance across such diverse settings suggests that large sequence models may exploit statistical regularities shared across subsets of environments despite substantial heterogeneity in observation modalities, action spaces, reward scales, and temporal horizons.
These results indicate that a single shared parameterization can achieve competitive performance across diverse settings, and suggest that large sequence models may benefit from statistical regularities shared across subsets of environments despite substantial heterogeneity in observation modalities, action spaces, reward scales, and temporal horizons.

% \begin{center}
% \end{center}

\subsection{Model Scaling Experiments}

\begin{figure}[h]
\centering
\includegraphics[width=0.7\textwidth]{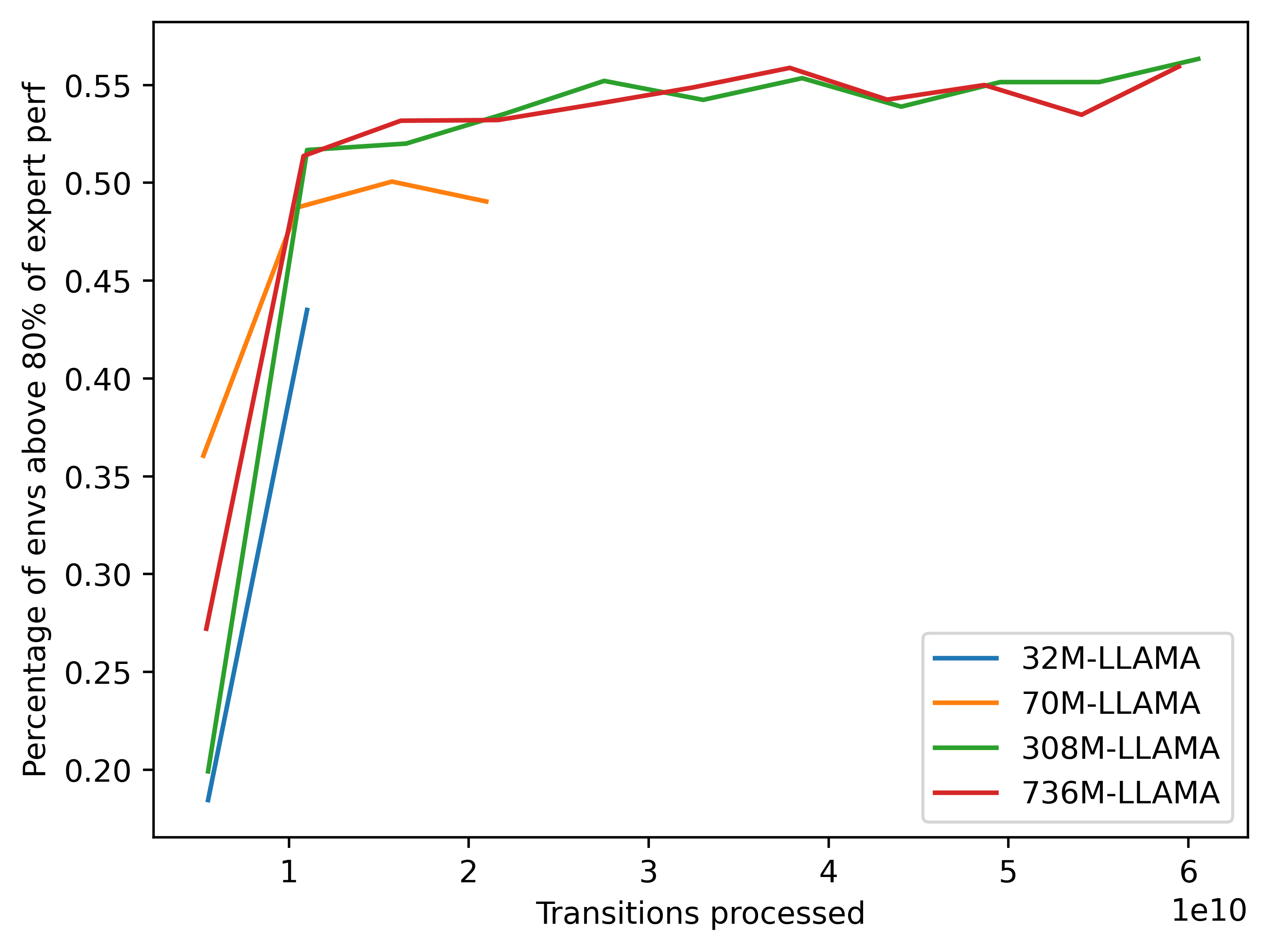}
\caption{\textbf{Model size scaling law}: In-distribution performance as a function of transitions processed.}
\label{fig:scalinglaw}
\end{figure}

Figure~\ref{fig:scalinglaw} shows the in-distribution performance of LDM-v0 as a function of model size. We evaluate four model scales: 32M, 70M, 308M, and 736M parameters.

All models use the same transformer context length of 2048 transitions. The batch size is fixed to 1024 for all models except the 736M variant, which uses a batch size of 384 due to GPU memory limitations.

For each model, we report the percentage of training environments where the pretrained model achieves at least 80\% of reference performance as training progresses.

We observe substantial performance improvements when increasing model size from 32M to 308M parameters, with larger models generally achieving stronger and more stable performance across training. Performance gains appear to plateau between the 308M and 736M models, although additional experiments would be required to characterize scaling behavior more precisely.

These results nevertheless suggest that heterogeneous offline RL sequence modeling can benefit from increased model capacity, similarly to trends observed in other large-scale sequence modeling domains.

% \begin{center}
% \end{center}

\section{Discussion and Future Work}

% We presented LDM-v0, a large-scale sequence model trained on multi-domain reinforcement learning environments through a unified offline learning pipeline. Our results suggest that a single transformer policy can model a broad range of RL behaviors within a shared parameterization across highly diverse environments.
We presented LDM-v0, a large-scale transformer policy trained through a unified offline reinforcement learning pipeline built on automated environment orchestration and large-scale trajectory generation across highly diverse RL environments. Our results demonstrate the feasibility of scalable heterogeneous RL pretraining using a single shared model across multiple domains and modalities.

A primary direction for future work is scaling the size and diversity of the training dataset: due to computational limitations we trained only on roughly 19\% of our current environment pool, which can be further expanded.% of In the current experiments, due to computational limitations, training trajectories were collected from approximately 3,000 environments, representing roughly 19\% of the environments currently integrated into our infrastructure. We are also continuing to expand the automation capabilities of our environment orchestration framework in order to increase both the number and diversity of supported RL libraries and environments.

An important limitation of the current work is that evaluations are primarily conducted on environments contained within the training distribution. Understanding the extent to which LDM-v0 acquires transferable decision-making strategies rather than learning predominantly environment-specific behaviors, remains an important open question. Future work will investigate both in-context adaptation and offline/online finetuning approaches for improving generalization to unseen environments.

More broadly, we hope this work motivates further research on scalable and automated reinforcement learning systems, including future large-scale pretrained reinforcement learning models.

\newpage
\appendix
\section{Libraries}
\label{appendix:libs}
\begin{table}[h!]
\centering
\begin{adjustbox}{width=0.71\columnwidth,center}
\begin{tabular}{| c | c || c | c |}
\hline
\textit{Library}& \textit{Environments} & \textit{Library} & \textit{Environments} \\
\hline
\hline
2048-gym-ai-environment & 1 & gym\_pomdp & 2 \\
\hline
2d\_robot\_arm\_dqn & 1 & gym\_probabilistic & 4 \\
\hline
484-curiosity & 135 & gym\_raas & 2 \\
\hline
against\_env & 1 & gym\_ressim & 1 \\
\hline
ai\_course\_project & 134 & gym\_reversi & 1 \\
\hline
ai\_football & 4 & gym\_rlcourse & 6 \\
\hline
artificial-intelligence-that-can-see & 1 & gym\_routing & 2 \\
\hline
asteroidsenv & 51 & gym\_rubik & 1 \\
\hline
baba\_is\_gym & 1 & gym\_short\_corridor & 1 \\
\hline
bandits & 1 & gym\_sokoban\_wrapped & 123 \\
\hline
bcr\_work & 1 & gym\_spaceinvaders & 1 \\
\hline
brdiv & 236 & gym\_splt & 1 \\
\hline
building-energy-storage-simulation & 1 & gym\_stag\_hunt & 3 \\
\hline
coding\_challenge & 1 & gym\_super\_mario\_bros & 138 \\
\hline
continuoussafetygym & 52 & gym\_supplychain & 7 \\
\hline
csc498 & 2 & gym\_toy & 8 \\
\hline
custom\_envs & 4 & gym\_trading\_env & 199 \\
\hline
cymnasium & 1 & gym\_vizdoom & 19 \\
\hline
d4rl\_atari & 100 & gym\_walk & 19 \\
\hline
dacbench & 3 & gym\_wildfire & 1 \\
\hline
deeprl-ltl & 1 & gym\_windy & 5 \\
\hline
diffuser & 7 & gym\_woodoku & 1 \\
\hline
domain\_randomization & 1 & gym\_yagw & 1 \\
\hline
driftgym & 3 & gymnasium\_2048 & 1 \\
\hline
drone-gym & 2 & gymnasium\_env & 1 \\
\hline
dsaa & 743 & gymnasium\_quadruped & 2 \\
\hline
dtqn-gelu & 21 & gymnasium\_robotics & 152 \\
\hline
eightnumber & 1 & gymnasium\_trading & 1 \\
\hline
final\_version\_in\_2stage & 152 & highway-env & 325 \\
\hline
fishing-gym & 11 & homestri\_ur5e\_rl & 1 \\
\hline
football2d & 4 & honoursresearchsim & 8 \\
\hline
frc\_gym & 1 & hurry\_taxi & 1 \\
\hline
growspace & 7 & hyphi\_gym & 1 \\
\hline
gym-bandits & 2 & icrl & 1 \\
\hline
gym-dubins-ac & 3 & indoor\_explorers & 2 \\
\hline
gym-game2048 & 99 & jogodavelhaia & 1 \\
\hline
gym-mobile-env & 3 & jsbgym & 55 \\
\hline
gym\_2048 & 1 & jumping-task & 1 \\
\hline
gym\_ai\_lab & 1 & lanro-gym & 4781 \\
\hline
gym\_antytrading & 2 & learning\_event\_triggered\_control & 947 \\
\hline
gym\_apl & 1 & macc & 943 \\
\hline
gym\_attitudecontrol & 1 & manipulator\_mujoco & 2 \\
\hline
gym\_barm & 10 & memtest & 1 \\
\hline
gym\_blobble & 1 & multiple-gym & 1 \\
\hline
gym\_bomber & 1 & mygym & 4 \\
\hline
gym\_classics & 12 & ogbench & 9 \\
\hline
gym\_cms & 1 & panda-gym & 19 \\
\hline
gym\_cog\_ml\_tasks & 9 & pogema & 1 \\
\hline
gym\_continuous-maze & 3 & probabilistic\_robotics & 2 \\
\hline
gym\_copter & 1 & procgen & 16 \\
\hline
gym\_crane & 1 & props & 51 \\
\hline
gym\_duane & 1 & pyflyt & 4 \\
\hline
gym\_dummy & 1 & qwertyenv & 1 \\
\hline
gym\_electricmotors & 54 & racingenv & 52 \\
\hline
gym\_extensions & 940 & rangl & 2 \\
\hline
gym\_fetch & 934 & rili\_co-adaptation & 4 \\
\hline
gym\_futbol & 1 & rl\_bicycle & 1 \\
\hline
gym\_games & 1 & rl\_reach & 49 \\
\hline
gym\_gmazes & 2 & rubiks\_cube\_gym & 7 \\
\hline
gym\_gridworlds & 2 & saferl-envs & 18 \\
\hline
gym\_gs & 1 & safety-gymnasium & 552 \\
\hline
gym\_inventory & 1 & saver & 3 \\
\hline
gym\_lorenz & 1 & shimmy & 45 \\
\hline
gym\_masked\_atari & 955 & simplegrid-gym & 102 \\
\hline
gym\_match\_input\_continuous & 4 & snake\_gym & 95 \\
\hline
gym\_md & 48 & sneks & 42 \\
\hline
gym\_missile\_command & 30 & spwk\_agtech & 1 \\
\hline
gym\_mosquitoes & 1 & stable-baselines3 & 940 \\
\hline
gym\_mountain\_cliff & 1 & tinycarlo & 1 \\
\hline
gym\_moving\_dot & 4 & uniform-coins-environment & 1 \\
\hline
gym\_n\_back & 2 & vizdoom & 9 \\
\hline
gym\_oscillator & 1 & voxelgym2d & 1 \\
\hline
gym\_pcgrl & 36 & widowx-reacher & 25 \\
\hline
\end{tabular}
\end{adjustbox}
\caption{Github libraries currently available in our Environments API}
\label{table:envsapi}
\end{table}

\newpage

\section{Datasets}
\label{appendix:datasets}

\begin{table}[h!]
\centering
\begin{adjustbox}{width=\columnwidth,center}
\begin{tabular}{| c | c | c || c | c | c |}
\hline
\textit{Library} & \textit{Environments} & \textit{Transitions} & \textit{Library} & \textit{Environments} & \textit{Transitions} \\
\hline
\hline
2d\_robot\_arm\_dqn & 1 & 3M & gym\_pomdp & 2 & 9M \\
\hline
484-curiosity & 74 & 186M & gym\_probabilistic & 3 & 11M \\
\hline
ai\_course\_project & 75 & 180M & gym\_raas & 2 & 11M \\
\hline
ai\_football & 2 & 24M & gym\_rlcourse & 4 & 16M \\
\hline
artificial-intelligence-that-can-see & 1 & 5M & gym\_sokoban\_wrapped & 2 & 9M \\
\hline
asteroidsenv & 18 & 54M & gym\_stag\_hunt & 3 & 9M \\
\hline
bandits & 1 & 3M & gym\_super\_mario\_bros & 70 & 181M \\
\hline
bcr\_work & 1 & 3M & gym\_supplychain & 5 & 15M \\
\hline
building-energy-storage-simulation & 1 & 3M & gym\_toy & 8 & 27M \\
\hline
continuoussafetygym & 52 & 167M & gym\_trading\_env & 163 & 523M \\
\hline
csc498 & 2 & 5M & gym\_walk & 15 & 62M \\
\hline
custom\_envs & 4 & 18M & gym\_windy & 5 & 20M \\
\hline
cymnasium & 1 & 3M & gym\_yagw & 1 & 3M \\
\hline
d4rl\_atari & 201 & 561M & gymnasium\_quadruped & 2 & 6M \\
\hline
dacbench & 3 & 21M & gymnasium\_robotics & 39 & 121M \\
\hline
deeprl-ltl & 1 & 2M & highway-env & 135 & 578M \\
\hline
diffuser & 5 & 24M & honoursresearchsim & 8 & 25M \\
\hline
domain\_randomization & 1 & 3M & hyphi\_gym & 1 & 3M \\
\hline
driftgym & 3 & 12M & icrl & 1 & 5M \\
\hline
dsaa & 177 & 525M & jogodavelhaia & 1 & 3M \\
\hline
final\_version\_in\_2stage & 6 & 17M & jsbgym & 49 & 132M \\
\hline
fishing-gym & 9 & 41M & lanro-gym & 5 & 15M \\
\hline
football2d & 2 & 3M & learning\_event\_triggered\_control & 228 & 665M \\
\hline
growspace & 3 & 8M & macc & 234 & 713M \\
\hline
gym-mobile-env & 1 & 5M & multiple-gym & 1 & 5M \\
\hline
gym\_ai\_lab & 1 & 3M & mygym & 2 & 6M \\
\hline
gym\_barm & 5 & 16M & panda-gym & 14 & 50M \\
\hline
gym\_blobble & 1 & 3M & probabilistic\_robotics & 2 & 9M \\
\hline
gym\_bomber & 1 & 5M & procgen & 7 & 24M \\
\hline
gym\_classics & 10 & 33M & props & 48 & 178M \\
\hline
gym\_cms & 1 & 2M & pyflyt & 3 & 17M \\
\hline
gym\_cog\_ml\_tasks & 9 & 28M & racingenv & 26 & 189M \\
\hline
gym\_copter & 1 & 2M & rili\_co-adaptation & 3 & 7M \\
\hline
gym\_electricmotors & 48 & 175M & rl\_bicycle & 1 & 5M \\
\hline
gym\_extensions & 243 & 666M & rl\_reach & 25 & 85M \\
\hline
gym\_fetch & 208 & 588M & saferl-envs & 16 & 55M \\
\hline
gym\_gridworlds & 2 & 8M & safety-gymnasium & 66 & 184M \\
\hline
gym\_inventory & 1 & 5M & saver & 3 & 9M \\
\hline
gym\_lorenz & 1 & 2M & shimmy & 23 & 99M \\
\hline
gym\_masked\_atari & 196 & 542M & simplegrid-gym & 101 & 463M \\
\hline
gym\_match\_input\_continuous & 4 & 8M & sneks & 24 & 62M \\
\hline
gym\_md & 48 & 253M & spwk\_agtech & 1 & 3M \\
\hline
gym\_missile\_command & 29 & 123M & stable-baselines3 & 128 & 297M \\
\hline
gym\_mosquitoes & 1 & 3M & uniform-coins-environment & 1 & 6M \\
\hline
gym\_pcgrl & 6 & 26M & widowx-reacher & 16 & 44M \\
\hline

\end{tabular}
\end{adjustbox}
\caption{Datasets used to train LDM-v0}
\label{table:dataset}
\end{table}

\newpage
\section{Candidate algorithm/configuration pairs}
\label{appendix:rl_candidate_configs}

\begin{table}[ht]
\centering
\scriptsize
\setlength{\tabcolsep}{3pt}
\renewcommand{\arraystretch}{1.5}
\begin{tabularx}{\textwidth}{llX}
\toprule
ID & Algorithm & Non-default hyperparameters \\
\midrule
C1  & A2C   & \textit{None} \\
C2 & A2C   & \texttt{ent\_coef=0.01} \\
C3 & ARS   & \textit{None} \\
C4 & DDPG  & \textit{None} \\
C5 & DDPG  & \texttt{learning\_starts=10000}, \texttt{batch\_size=100}, \texttt{train\_freq=50}, \texttt{gradient\_steps=-1} \\
C6  & DDPG  & \texttt{learning\_starts=2000}, \texttt{batch\_size=64}, \texttt{train\_freq=50}, \texttt{gradient\_steps=-1}, \texttt{learning\_rate=1e-4}, \texttt{tau=0.01} \\
C7 & DQN   & \textit{None} \\
C8 & PPO   & \textit{None} \\
C9  & PPO   & \texttt{n\_steps=4000}, \texttt{n\_epochs=4}, \texttt{gae\_lambda=0.97}, \texttt{ent\_coef=0.01}, \texttt{batch\_size=512} \\
C10 & PPO   & \texttt{n\_epochs=4}, \texttt{batch\_size=512} \\
C11 & QR-DQN & \textit{None} \\
C12  & SAC   & \textit{None} \\
C13  & SAC   & \texttt{learning\_starts=10000}, \texttt{batch\_size=100}, \texttt{train\_freq=50}, \texttt{gradient\_steps=-1}, \texttt{learning\_rate=1e-3} \\
C14  & TD3   & \textit{None} \\
C15 & TD3   & \texttt{learning\_starts=10000}, \texttt{batch\_size=100}, \texttt{train\_freq=50}, \texttt{gradient\_steps=-1} \\
C16  & TQC   & \textit{None} \\
C17  & TRPO  & \textit{None} \\
C18  & TRPO  & \texttt{n\_steps=4000}, \texttt{n\_critic\_updates=80}, \texttt{gae\_lambda=0.97}, \texttt{cg\_max\_steps=10} \\
C19 & TRPO  & \texttt{n\_steps=1024}, \texttt{target\_kl=0.001}, \texttt{cg\_max\_steps=10}, \texttt{gae\_lambda=1.0}, \texttt{cg\_damping=0.01}, \texttt{learning\_rate=3e-4}, \texttt{n\_critic\_updates=3} \\
\bottomrule
\end{tabularx}
\caption{
Candidate reinforcement learning algorithm/configuration pairs used for automated reference-policy generation. \textit{None} indicates that no non-default hyperparameters were used, i.e., the corresponding Stable-Baselines3 or SB3-Contrib library defaults are used.  Candidate configurations are used only when compatible with the environment observation and action spaces.
}
\label{tab:rl_candidate_configs}
\end{table}

\section*{Acknowledgments}
We thank the NeoInstinct developers for their valuable technical contributions to the development of our environment orchestration framework and to the collection and integration of experimental environments. Their work was essential to the successful development and execution of this study. We also thank our NeoInstinct colleagues for helpful discussions that contributed to the development of this work.

This work was funded by NeoInstinct SA.
%Dawid Malysa, Dominik Heinisch, Krzysztof Matuszewski, Ryszard Leykam, Marek Kostykowski, and Mariusz Karpowicz
%Anthony Fillion, Flavio Mor and Fabio Cesa

%%%%%%%%%%%%%%%%%%%%%%%%%%%%%%%%%%%%%%%%%%%%%%%%%%%%%%%%%%%%%%%%
%% NOTE: THIS MARKS THE END OF THE "MAIN TEXT"
%%%%%%%%%%%%%%%%%%%%%%%%%%%%%%%%%%%%%%%%%%%%%%%%%%%%%%%%%%%%%%%%

%%%%%%%%%%%%%%%%%%%%%%%%%%%%%%%%%%%%%%%%%%%%%%%%%%%%%%%%%%%%%%%%
%% Bibliography
%%%%%%%%%%%%%%%%%%%%%%%%%%%%%%%%%%%%%%%%%%%%%%%%%%%%%%%%%%%%%%%%
% \newpage
\bibliography{main}
\bibliographystyle{rlj}

%%%%%%%%%%%%%%%%%%%%%%%%%%%%%%%%%%%%%%%%%%%%%%%%%%%%%%%%%%%%%%%%
% AUTHOR: If your paper has no supplementary materials, you may 
%         comment out the line below, which creates the title for
%         the supplementary materials.
%%%%%%%%%%%%%%%%%%%%%%%%%%%%%%%%%%%%%%%%%%%%%%%%%%%%%%%%%%%%%%%%
% \beginSupplementaryMaterials

% Content that appears after the references are not part of the ``main text,'' have no page limits, are not necessarily reviewed, and should not contain any claims or material central to the paper. 
% %
% If your paper includes supplementary materials, use the \begin{center}
%     {\tt {\textbackslash}beginSupplementaryMaterials} 
% \end{center}
% command as in this example, which produces the title and disclaimer above. 
% %
% If your paper does not include supplementary materials, this command can be removed or commented out.

\end{document}